\documentclass{article}
\usepackage{iclr2027_conference,times}
\usepackage{float,needspace}
\usepackage{hyperref}
\usepackage{url}
\usepackage{flafter}
\usepackage{placeins}
\usepackage{graphicx}
\usepackage{booktabs}
\usepackage{longtable}
\usepackage{array}
\usepackage{amsmath,amssymb}
\usepackage{xcolor}
\definecolor{citecolor}{HTML}{2A6F8F}
\definecolor{figrefcolor}{HTML}{A23B5A}
\hypersetup{
    colorlinks=true,
    citecolor=citecolor,
    linkcolor=figrefcolor,
    urlcolor=citecolor,
    pdftitle={More Programs or More Rolls? Separating Coverage from Specialization in LLM Harnesses},
    pdfauthor={Ziyang Xu, Haitian Zhong, Hao Zhou, Hao Qin, Chenhan Jin, Te Qi, Shengze Xu, Tieyong Zeng}
}

\iclrfinalcopy  

\title{More Programs or More Rolls?\\Separating Coverage from Specialization in LLM Harnesses\thanks{Code is available at: \url{https://github.com/StatXzy7/harness-eval}.}}

\author{Ziyang Xu\textsuperscript{1}, Haitian Zhong\textsuperscript{2,3},
Hao Zhou\textsuperscript{1}, Hao Qin\textsuperscript{1} \\
\textbf{Chenhan Jin\textsuperscript{1}, Te Qi\textsuperscript{4},
Shengze Xu\textsuperscript{1}, Tieyong Zeng\textsuperscript{5,6}\thanks{Corresponding author.}} \\
\textsuperscript{1}Department of Mathematics, The Chinese University of Hong Kong \\
Hong Kong, China \\
\textsuperscript{2}New Laboratory of Pattern Recognition (NLPR), \\
State Key Laboratory of Multimodal Artificial Intelligence Systems (MAIS), \\
Institute of Automation, Chinese Academy of Sciences \\
\textsuperscript{3}Zhongguancun Academy \\
\textsuperscript{4}Department of Basic Research, \\
SINOPEC Research Institute of Petroleum Processing \\
\textsuperscript{5}Institute for Advanced Study, \\
Beijing Normal-Hong Kong Baptist University, Zhuhai, China \\
\textsuperscript{6}School of Mathematics and Statistics, Guangzhou Nanfang College \\
Guangzhou, China \\
\texttt{ziyang.xu@link.cuhk.edu.hk}, \texttt{tieyongzeng@bnbu.edu.cn}
}

\newcommand{\bare}{single-call}
\newcommand{\react}{\texttt{react}}

\newcommand{\RevisionPrimaryDelta}{-0.16}
\newcommand{\RevisionPrimaryCI}{[-0.83,\ +0.87]}
\newcommand{\RevisionPrimaryP}{0.647}
\newcommand{\RevisionPrimaryRange}{-4.36\text{ to }+2.74}
\newcommand{\RevisionGateDelta}{-0.69}

\newcommand{\RevisionKmatchedDelta}{-0.31}

\newcommand{\RevisionRtwoCached}{+0.42}
\newcommand{\RevisionRtwoOff}{+0.81}
\newcommand{\RevisionRtwoChange}{+0.39}
\newcommand{\RevisionRtwoChangeCI}{[-0.95,\ +1.63]}
\newcommand{\RevisionRtwoFlips}{2391}
\newcommand{\RevisionRtwoFlipPercent}{8.54}
\newcommand{\RevisionRtwoRepeatFlips}{1245}
\newcommand{\RevisionRtwoRepeatPercent}{8.65}

\newcommand{\RevisionCostRows}{153{,}600}
\newcommand{\RevisionLogicalCalls}{259{,}304}

\newcommand{\RevisionMathDisagreement}{10.6}
\newcommand{\RevisionMathOracle}{98.50}
\newcommand{\RevisionMathBest}{93.75}
\newcommand{\RevisionMathHeadroom}{4.75}

\newcommand{\RcvPairedN}{386}

\newcommand{\RcvCells}{10{,}422}
\newcommand{\RcvSingleCells}{13{,}896}
\newcommand{\RcvExcludedN}{14}
\newcommand{\RcvOriginalMissingCells}{61}
\newcommand{\RcvOriginalExcludedPct}{3.50}
\newcommand{\RcvRealPlugin}{2.33}
\newcommand{\RcvClonePlugin}{2.16}
\newcommand{\RcvRealG}{0.10}
\newcommand{\RcvCloneG}{0.35}
\newcommand{\RcvD}{-0.26}
\newcommand{\RcvDCI}{[-1.45, 0.95]}
\newcommand{\RcvP}{0.8127}
\newcommand{\RcvRealRho}{0.789}
\newcommand{\RcvCloneRho}{0.006}
\newcommand{\RcvRealRhoCI}{[0.75, 0.83]}
\newcommand{\RcvCloneRhoCI}{[-0.07, 0.08]}
\newcommand{\RcvRhoDiff}{0.783}
\newcommand{\RcvRhoDiffCI}{[0.69, 0.87]}
\newcommand{\RcvBaselineAcc}{95.85}
\newcommand{\RcvPolicyGain}{0.00}
\newcommand{\RcvConflictN}{383}
\newcommand{\RcvConflictD}{-0.61}
\newcommand{\RcvConflictDCI}{[-1.96, 0.64]}

\newcommand{\RcvReplayFullCoverage}{98.70}
\newcommand{\RcvLossPairs}{570}
\newcommand{\RcvLossTasks}{100}
\newcommand{\MathUnique}{29}
\newcommand{\MathGenUnique}{28}
\newcommand{\MathGenMin}{69.69}
\newcommand{\MathGenMax}{92.49}
\newcommand{\MathMaxDisagreement}{33.16}
\newcommand{\MathIdenticalPairs}{2.22}
\newcommand{\MathUnionRepair}{84.21}
\newcommand{\MathUnionFail}{0.82}
\newcommand{\RcvReplayThreeGap}{+0.20}
\newcommand{\RcvReplayNineGap}{+0.08}
\newcommand{\RcvReplayFifteenGap}{+0.02}

\renewcommand{\RevisionMathDisagreement}{10.94}
\renewcommand{\RevisionMathOracle}{99.22}
\renewcommand{\RevisionMathBest}{95.08}
\renewcommand{\RevisionMathHeadroom}{4.15}

\begin{document}
\maketitle
\lhead{Preprint}

\begin{abstract}
Automated generation of LLM harnesses promises to improve inference
through task specialization. Yet additional answer coverage can arise
from repeated execution of the same program, making specialization
difficult to identify. We introduce a controlled evaluation that
separates answer coverage, repeatable task advantages, and gains from
pre-execution selection. On $\RcvPairedN$ MATH-500 tasks, we compare
eight generated harnesses plus a baseline with nine byte-identical
baseline copies, using three executions per member. Identical programs
yield $\RcvClonePlugin$ percentage points of repeat-averaged oracle
headroom. Generated programs exhibit substantially more repeatable
score patterns, but these chiefly reveal persistent weaknesses:
losses relative to the baseline persist across all three repeats on
$\RcvLossTasks$ tasks, while persistent wins occur on only one task
and are sensitive to answer extraction.
The frozen selector gains $\RcvPolicyGain$ percentage points, and both
populations reach $\RcvReplayFullCoverage\%$ oracle coverage at
$27$ harness executions. Stable complementarity remains unresolved
at three repeats. Supporting BIRD traces locate failures in mechanism
implementation, activation, and output validity. Together, these
findings establish why coverage and repeatability alone cannot justify
claims of useful specialization. They motivate an evaluation standard
for harness diversity: task advantages should persist across executions,
guide usable decisions, and improve on additional fixed-program
executions under matched inference budgets.
\end{abstract}

\section{Introduction}\label{sec:introduction}
Automated agent design allows language models to write the programs that
organize a solver's inference. These LLM harnesses arrange prompts, tools,
and control flow around a fixed solver
\citep{adas2024,tthe2026,selfharness2026}. A population of such programs
offers an attractive possibility: a division of labor in which different
members address different task-specific failures. Realizing this
possibility would let a system allocate inference to the program best
suited to each task. It raises a concrete question: when should a fixed
budget buy more programs, and when would further executions of one
program suffice?

The most immediate evidence for this division of labor is also
ambiguous. A population can contain more correct answers than any
single member simply because it provides more attempts. Pass@$k$ measures the probability of at least one correct answer
among $k$ samples; self-consistency and
sampling-and-voting ensembles turn repeated attempts into an aggregated
answer \citep{chen2021codex,selfconsistency2022,moreagents2024}.
An oracle that selects after observing correctness can exploit
complementary successes even when every member has identical code.
Additional coverage therefore leaves open how much the population
benefits from having different programs.

Repeating the evaluation introduces a second distinction. With few
repeats, oracle estimates can still select favorable estimation errors;
even strongly repeatable score patterns can reflect a program's
persistent weaknesses. Exploiting specialization before execution requires task advantages
that recur and can be recognized from the task.
We therefore distinguish \emph{answer coverage},
\emph{repeat-stable task advantages}, and \emph{usable selection}.
These distinctions turn the promise of program specialization into
an evaluation problem: identify which advantages persist, which can
guide a choice, and what they add beyond repeated execution of a
fixed program.

\begin{figure}[t]
\centering
\includegraphics[width=\linewidth]{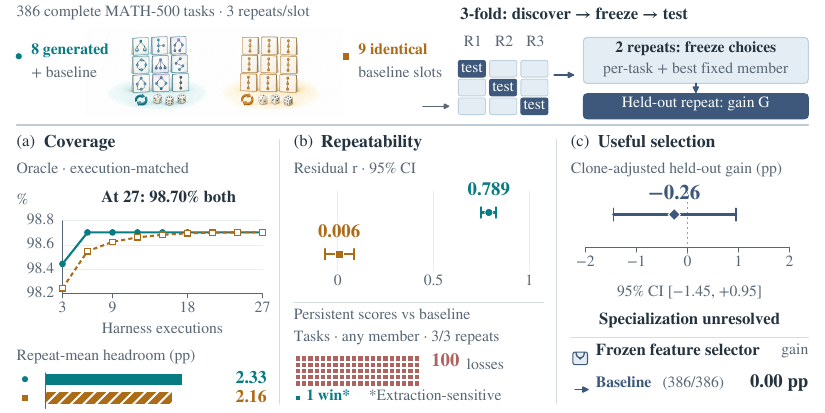}
\caption{\textbf{Coverage, repeatability, and selection on $\RcvPairedN$ complete
MATH tasks.} Oracle coverage matches executions, not tokens/calls; headroom
uses repeat means. Persistent task counts require at least one generated
member to win/lose against baseline in all three repeats; the sole win is
extraction-sensitive. Here $G$ is held-out task-wise selection gain over the discovery-selected fixed member; panel (b) labels residual Pearson correlation as $r$. R1--R3 are repeat labels; pp denotes percentage points and CI denotes confidence interval. Intervals are 95\% bootstrap CIs. Program drawings
are schematic.}
\label{fig:sampling-control}
\end{figure}

We investigate this problem on MATH-500, starting with the coverage
offered by a generated population. We then compare a selected low-call
panel with byte-identical baseline copies, holding the solver,
population size, and repeat count fixed. Choices learned from earlier
executions are tested on held-out repeats. A selector frozen on
development data tests pre-execution choice, and oracle replay compares
coverage at equal harness execution counts. Separate BIRD traces
examine whether proposed mechanisms are implemented, triggered, and
able to return valid answers. This sequence follows the evidence needed
to move from additional answers to useful specialization.

The central empirical finding is that the generated panel's repeatable
differences predominantly identify weaknesses
(Figure~\ref{fig:sampling-control}). Persistent losses span
$\RcvLossTasks$ tasks, while persistent wins occur on one task and are
sensitive to answer extraction. Identical programs already yield
positive oracle headroom. The clone-adjusted held-out gain is
$\RcvD$ pp, leaving stable complementarity unresolved at three repeats.
The frozen selector gains $\RcvPolicyGain$ pp, and both arms reach
$\RcvReplayFullCoverage\%$ oracle coverage at $27$ executions.
Thus, substantial repeatability coexists with zero gain for the tested
selector and matching full-budget coverage from further baseline
executions.

The study provides a controlled basis for attributing population gains:
a same-code sampling reference, a directional analysis of repeated wins
and losses, and separate evaluations of pre-execution choice and
post-execution coverage. Together with the execution traces, these
analyses connect program differences to the behavior and decisions
through which they could become useful. The resulting design target is
a task advantage that persists, guides a usable decision, and adds value
beyond further attempts with a fixed program.

\section{Related Work}\label{sec:related}
\paragraph{Automated agent and harness design.}
Executable programs provide a flexible space for adapting LLM behavior.
ADAS searches this space directly \citep{adas2024}; Self-Harness uses
failure traces to propose edits and regression tests to retain them
\citep{selfharness2026}; TTHE adapts the harness during evaluation from
unlabeled traces \citep{tthe2026}. Related systems improve task
adaptation and validation allocation \citep{jitagent2026,coevolve2026},
or expose harness behavior through verification, diagnosis, localization,
and policy evolution
\citep{harnesslens2026,harnessfix2026,ahe2026,gatedqd2026,handbook2026,harnessforge2026}.
\citet{metaagents2025} report low diversity in task-level score patterns
among the automatically designed agents they study. We examine program
populations through both execution traces and repeated outcomes:
traces reveal which mechanisms execute, and same-code repeats measure
coverage from sampling. This perspective connects program equivalence
and mutation testing \citep{budd1982,papadakis2019} with performance
attribution in generated harness populations.

\paragraph{Repeated sampling and ensembles.}
Multiple executions can improve answer coverage even with a fixed
program. Pass@$k$ measures success within $k$ samples; self-consistency
and identical-agent ensembles aggregate attempts
\citep{chen2021codex,selfconsistency2022,moreagents2024}, while nominally
temperature-$0$ calls can also vary \citep{nondeterminism2023}. These
gains provide the sampling reference in our execution-matched
comparison. A second distinction concerns when selection occurs.
Post-execution verifiers rank candidate solutions \citep{llmverifier2026};
ensemble selection chooses a model combination on validation data and
averages predictions at inference \citep{caruana2004}. Our frozen
selector chooses a program before execution, while oracle replay
measures coverage across recorded outputs.

\paragraph{Selection and resource allocation.}
Turning a population into a useful choice connects harness design to
algorithm selection and portfolios
\citep{rice1976,xu2008satzilla,kotthoff2016}. HELIX uses complementary
sibling trajectories to expand verified coverage and supply data for
model improvement \citep{helix2026}. STS compares six reasoning
paradigms and trains an embedding-based router to recover part of
their oracle gap \citep{sts2026}. Harness sensitivity across
model--problem pairs \citep{nouniversal2026} and evaluated models
\citep{nonmonotone2026}, search routing \citep{cfrouting2026}, and
preference-trained model routing \citep{routellm2024} further motivate
task-dependent choices. GRASP selects edits to a bounded skill library
under regression constraints \citep{grasp2026}. We hold the target model
fixed and measure oracle coverage, repeat-stable task advantages, and
realized selector gain in the same program population. The distinction
between quantitative and qualitative interaction \citep{gailsimon1985}
clarifies why varying effect sizes can coexist with an unchanged best
action.

\section{Problem Formulation}\label{sec:measures}
We distinguish the answers a population provides, the task advantages
that persist across executions, and the gains a selector realizes.
Let $P$ be a population of harness members containing the single-call
baseline $H_0$. Write $h\in P$ for a member, $x_i$ for task $i$ among
$n$ evaluated tasks, and $r\in\{1,\ldots,R\}$ for one of $R$ execution
repeats per member--task pair. The score $Y(h,x,r)\in\{0,1\}$ is one
exactly when member $h$ answers task $x$ correctly in repeat $r$.
Equations use proportions; multiplying by $100$ gives accuracies in
\% and gains or headroom in percentage points (pp).
Appendix~\ref{app:notation} summarizes the notation.

\paragraph{Coverage: which additional answers are available?}
A one-run oracle selects a member after seeing correctness. For a fixed repeat $r$, its
headroom over the best fixed member in the recorded matrix is
\begin{equation}
H(P)=\tfrac1n\sum_{i=1}^n\max_{h\in P}Y(h,x_i,r)
-\max_{h\in P}\tfrac1n\sum_{i=1}^nY(h,x_i,r).
\label{eq:headroom}
\end{equation}
This quantity captures available correct answers, including those
obtained through repeated sampling. The repeat-averaged plugin
$\widehat H$ replaces $Y(h,x,r)$ in Equation~\ref{eq:headroom} with
$\hat p_h(x)=R^{-1}\sum_{r=1}^R Y(h,x,r)$ on complete observations.
With few repeats,
maximizing those means can select favorable estimation errors,
which motivates the same-code control.

\paragraph{Stable complementarity: which member is best for each task?}
Write $p_h(x)=\Pr\{Y(h,x,r)=1\}$ for member $h$'s success probability
over fresh executions under a fixed execution condition.
The corresponding population headroom is
$H_{\rm stable}=\mathbb E_x\max_h p_h(x)-\max_h\mathbb E_x p_h(x)$,
where $\mathbb E_x$ averages over the target task distribution and
all member maxima range over $P$.
For $P=\{H_0,H_1\}$, abbreviate $p_{H_j}(x)$ as $p_j(x)$ for
$j\in\{0,1\}$ and define $\Delta(x)=p_1(x)-p_0(x)$. Then
\begin{equation}
H_{\rm stable}=\tfrac12\bigl(\mathbb E_x|\Delta(x)|-|\mathbb E_x\Delta(x)|\bigr).
\label{eq:identity}
\end{equation}
Positive complementarity requires task-wise crossover: the better
member changes with the task. If one member is better everywhere,
headroom is zero even when its advantage varies in size. Identical
programs under the same condition have equal $p_h(x)$, giving a
zero-complementarity reference. This definition concerns population
probabilities; the repeat diagnostic below tests choices learned from
finite executions.

For a concrete example, consider two equally common task types with
success probabilities $(0.90,0.60)$ for $H_0$ and $(0.80,0.20)$ for
$H_1$. The performance gap is task-dependent, but $H_0$ remains
preferable on both types. The task-wise oracle and the best fixed
member therefore both achieve $0.75$, leaving zero stable headroom.
This constructed example separates variation in the size of an
advantage from a change in the preferred action. It motivates checking
the direction of repeated differences, alongside their repeatability.

\paragraph{Usable selection: which advantage can guide a decision?}
A selector $\pi_Z$ maps the pre-execution features $Z(x)$ of task $x$
to a member $\pi_Z(x)\in P$.
Its usefulness depends on whether those features predict the relevant
advantage. We evaluate this decision separately from choices learned
from earlier outcomes on the same task, which test repeat
generalization. An execution-matched oracle replay then compares the
coverage offered by a portfolio with additional fixed-program
executions. Its budget counts harness executions; each execution may
contain multiple model calls and tokens.

\section{Experimental Design}\label{sec:setup}
\subsection{Harness Populations and Evaluation Setup}
\label{sec:interface}
We hold the target solver, GLM-5.3-Flash, fixed and vary the
Python programs that call it. A \emph{builder} writes each harness,
the \emph{solver} answers its model calls, and an external \emph{judge}
scores the final output. Gold answers are available only to the judge.
Each harness receives a question and can query the fixed solver.
BIRD harnesses also have access to database schemas and SQL execution
feedback. The \bare{} baseline makes one temperature-$0$
call. Generated programs can change prompts, decoding, sample count,
and control flow; interface and model details appear in
Appendix~\ref{app:inventory}.

MATH-500 tests whether additional coverage survives repetition and
supports useful program choices. Table~\ref{tab:study-map} maps these
questions to their comparisons. BIRD separately examines how proposed
mechanisms execute and what admission screening changes.

\begin{table}[htbp]
\centering\small
\setlength{\tabcolsep}{4pt}
\begin{tabular*}{\linewidth}{@{}l@{\extracolsep{\fill}}ll@{}}
\toprule
Question & Key comparison & Readout \\
\midrule
\textbf{Additional coverage} & Full pool vs. best fixed & Oracle headroom $H(P)$ \\
\addlinespace[3pt]
\textbf{Repeat transfer} & Generated vs. same-code & Held-out gain $D$ \\
\addlinespace[3pt]
\textbf{Usable selection} & Frozen selector vs. dev-fixed & Selection gain $S(\pi_Z)$ \\
\addlinespace[3pt]
\textbf{Budgeted coverage} & Portfolio vs. repeated baseline & Oracle coverage \\
\bottomrule
\end{tabular*}
\caption{\textbf{Comparisons separating coverage from usable specialization.}
Oracle comparisons use correctness after execution; the frozen selector
chooses from the question before execution. \mbox{Dev-fixed} denotes the
development-selected fixed member. $D$ is generated-panel minus clone
held-out selection gain (Equation~\ref{eq:clone-difference}); $S(\pi_Z)$ is
the frozen selector's accuracy gain over dev-fixed.}
\label{tab:study-map}
\end{table}

\paragraph{MATH tasks and scoring.}
\label{sec:math-setup}
MATH-500 \citep{math500} is split into $100$ development and $400$
evaluation problems. All primary MATH comparisons use the same
$\RcvPairedN$ evaluation problems, selected after collection by requiring
complete observations in both repeat arms and the single-run population.
This excludes $\RcvExcludedN$ problems with incomplete repeat observations;
Appendix~\ref{app:wp1r} gives completeness counts, exclusions, and
conflict sensitivity. The judge extracts the
final answer, preferring the designated final-answer line, and
compares numeric values when both parse or normalized LaTeX otherwise
(Appendix~\ref{app:crossdomain}).

\paragraph{MATH generation and repeat panel.}
Six builders---GLM-5.3, Qwen3.8-Max, DeepSeek-V4-Pro, Kimi-K3,
MiniMax-M3, and ERNIE-5.0-Thinking-Preview---each receive eight slots
under generation seed $0$. A slot is an independent opportunity to improve on a
single greedy call: the builder chooses the strategy, with no cross-slot
diversity constraint. Each slot allows up to three attempts at
temperature $0.7$ and $32{,}768$ output tokens, using the same prompt
without feedback from earlier attempts. Generation stops at the first
program that runs successfully and returns nonempty answers on two
development questions. The $48$ slots yield $35$ frozen
programs (builder instructions: Appendix~\ref{app:builder-instruction}).

The repeat panel samples eight of these programs. A historical mean-call
ceiling of $3.05$ leaves $21$ eligible members; a fixed salted hash
selects one per builder and two at-large seats. The resulting low-call
panel includes draft--check--revise, agreement checking, and sample
voting (Appendix~\ref{app:panel-mechanisms}). Adding \bare{} gives
nine members. This panel and nine identical baseline slots each receive
three fresh, cache-disabled repeats, with decoding fixed by the source.

The selected programs differ in how they obtain and resolve additional
answers. Some request several samples and vote on normalized answers;
others verify a draft, obtain an alternative solution, or request
adjudication when answers disagree. One uses differently prompted
solutions and stops when three agree. Verification and adjudication
are additional prompts to the same solver, without access to gold
answers. A single logical call can request five samples, and the
$3.05$ eligibility threshold is a historical mean rather than a
per-task maximum. The inventory describes implemented operations;
their presence alone does not show that each branch activates or
improves an answer.

\paragraph{BIRD discovery and admission.}
\label{sec:bird-setup}
BIRD \citep{bird2021} requires SQL for a question and database. In the
discovery study, TTHE uses a GLM-5.3-Flash builder to produce
$12$ candidates that pass basic execution checks. These
programs and two supplied baselines are evaluated on $60$ questions.
The admission study crosses free versus assigned (``forced'') strategies with
ungated versus conformance-gated admission: six builders, three seeds,
and eight slots per builder--seed cell in each arm, with up to three attempts per slot
(temperature $0.7$; $16{,}384$ output tokens). Forced strategies
include execution repair, three-sample voting, and schema linking.
All arms require executable programs and nonempty SQL; gated arms also
check mechanism conformance and exclude some structures.

The admission split uses $365$ questions from two databases for
development and $1{,}169$ from nine databases for evaluation, with a
pre-drawn $400$-question core shared by all four arms. An $18$-item
smoke test is a development-exposure exception. The official set-of-rows
scorer compares predicted and gold query results, with a $30$-second
timeout and failures scored zero. Discovery uses an earlier
stringified-row proxy with a row cap. Admission uses shared response
caching, whereas the MATH repeats disable caching. Full split, scoring,
and bundled gating specifications appear in
Appendices~\ref{app:phase1-provenance} and~\ref{app:phase2-protocol}.

\subsection{Same-Code Controls and Repeat Validation}\label{sec:protocol}
\paragraph{Learning a choice on two repeats, testing it on the third.}
\label{sec:calibration-test}
Each fold uses two discovery repeats to select a member for every task
and a single best fixed member. The third repeat measures their accuracy
difference. Averaging over tasks and the three folds gives gain $G$: $G_{\rm real}$ for the generated panel plus baseline and $G_{\rm clone}$ for the same-code arm. We compare paired task-level gains on the common
complete tasks through
\begin{equation}
D=G_{\rm real}-G_{\rm clone}.
\label{eq:clone-difference}
\end{equation}
A positive difference means that choices learned from past outcomes
transfer better for distinct programs than for identical ones.
Both the task-wise choice and the fixed comparator are determined
entirely by the discovery repeats.
Thus, $G$ evaluates a choice learned from finite observations: it depends
on both the underlying task advantages and how accurately the discovery
repeats identify them. $D$ compares this learning procedure across the
two arms. It is distinct from $H_{\rm stable}$, which assumes access to
the success probabilities themselves. Its confidence interval is
therefore not an interval for $H_{\rm stable}$; a near-zero $D$ need
not imply a small population-level opportunity.

Positive support requires $D>1.0$ pp, positive 95\% bootstrap lower
bounds for both $D$ and $G_{\rm real}$, and randomization $p<0.05$.
Before subsetting, incomplete tasks must account for at most $10\%$
of either arm or their paired intersection, with at least $30$ paired
tasks retained. These conditions combine effect size, uncertainty,
and observation completeness; failure to satisfy them yields
\textsc{Insufficient Evidence}.

\paragraph{Distinguishing repeatability from advantage.}
We denote by $\rho_{\rm res}$ the mean held-out-versus-discovery Pearson correlation of score residuals across repeats, after removing additive member and task effects. This exploratory
measure describes repeatability, while $G$ describes held-out selection
gain. Known-truth simulations characterize the diagnostic's sensitivity.
Appendices~\ref{app:repeatability} and~\ref{app:statistics} specify these
calculations, tie handling, uncertainty estimates, and decision rules.
The acquisition protocol preceded collection; the repeat diagnostic and
common-complete-set analysis were developed afterward. Recovery retained
the programs, solver, tasks, and repeat design (Appendix~\ref{app:wp1r}).

\paragraph{Checking the diagnostic against known truth.}
Calibration includes equal-ability and global-dominance settings with
$H_{\rm stable}=0$, alongside alternatives in which the best member
changes with the task. In the strong two-specialist case, $400$ tasks
form two equal halves. Each specialist succeeds with probability $0.95$
on its own half and $0.60$ on the other; the remaining seven members
have probability $0.60$ throughout. The oracle accuracy is $0.95$ and
the best fixed accuracy is $0.775$, giving $17.5$ pp of true headroom.
These settings distinguish false support from failure to detect an
existing advantage. The simulation variant uses fewer resampling draws
and checks completeness in the real arm; the empirical rule also checks
clones and the paired intersection. Appendix~\ref{app:statistics}
specifies both rules and their tie handling.

\subsection{Frozen Selection and Execution-Matched Replay}\label{sec:selection-design}
\paragraph{Selecting a program from the question.}
Using only development data, we fit $\pi_Z$ to predict the member
with the highest available-repeat mean correctness on each task;
training-label ties follow member-ID order. Five task folds keep each
task's repeats together. The fitted models share a term-frequency--inverse-document-frequency (TF-IDF) vocabulary
learned from development questions and use training-fold member
accuracies as additional inputs. At inference, we average the fitted
fold models' probability vectors and select the top member, falling
back to \bare{} when the top-two gap is below $0.15$, including a tie
for the highest probability. The policy is frozen before evaluation;
Appendix~\ref{app:selection} specifies fitting and fallback details.
Comparators include the development-selected fixed member $h_{\rm dev\text{-}fixed}$ and one randomly drawn fixed member. We write $S(\pi_Z)$ for the policy accuracy minus that of $h_{\rm dev\text{-}fixed}$, averaging over evaluation tasks and repeats.

\paragraph{Allocating executions to a portfolio.}
An offline oracle replay compares a development-ranked portfolio with
same-code executions at per-task budgets $b\in\{3,9,15,27\}$ harness executions. Each included
member contributes three repeats. For clones, we compute exact mean
coverage over uniform size-$b$ subsets of the $27$ recorded executions.
Both arms use post-execution correctness to measure available answers.
The recorded selector choices and portfolio order stay fixed for the
recovery analysis. The budget unit is a harness execution, with model-call and token use determined
by the program. Appendix~\ref{app:selection} gives the policy, replay
formula, and accounting details.

\section{Results}\label{sec:results}\label{sec:diagnosis}
\subsection{The Population Covers More Tasks Than Any Single Member}
\label{sec:f2-headroom}
The full generated population plus \bare{} reaches
$\RevisionMathOracle\%$ oracle coverage, compared with
$\RevisionMathBest\%$ for the best fixed member, \bare{}.
Every generated member is individually worse
($\MathGenMin$--$\MathGenMax\%$), yet their union adds
$\RevisionMathHeadroom$ pp of coverage available to an oracle
after execution.

This coverage comes with substantial variation in scored outcomes:
$\MathUnique$ of the $36$ correctness vectors are unique, mean
pairwise disagreement is $\RevisionMathDisagreement\%$
(maximum $\MathMaxDisagreement\%$), and the union repairs
$\MathUnionRepair\%$ of baseline errors. The next question is how
much of this variation survives repetition. The following comparisons
use the selected panel and its fresh executions, with baseline accuracy
$\RcvBaselineAcc\%$, to distinguish repeatable advantages from
sampling. Their population and executions differ from the full-pool
single-run coverage reported here (\S\ref{sec:math-setup}).

\subsection{Identical Programs Also Produce Positive Oracle Estimates}
\label{sec:f3-clones}
Identical programs produce a $\RcvClonePlugin$ pp repeat-averaged
oracle estimate, compared with $\RcvRealPlugin$ pp for the generated
panel (Table~\ref{tab:clone-control}). The nine clone slots implement
the same program, so their population-level stable complementarity
is zero by construction. Thus, even after averaging three repeats,
a positive oracle estimate
alone does not identify specialization.

\begin{table}[htbp]
\centering\small
\begin{tabular}{@{}lcc@{}}
\toprule
Measure & Generated + \bare{} & Same-code clones \\
\midrule
Oracle headroom $\widehat H$ (pp) & $\RcvRealPlugin$ & $\RcvClonePlugin$ \\
Residual correlation $\rho_{\rm res}$ & $\RcvRealRho$ & $\RcvCloneRho$ \\
Held-out selection gain $G$ (pp) & $\RcvRealG$ & $\RcvCloneG$ \\
\midrule
\multicolumn{3}{l}{Clone-adjusted gain: $D=\RcvD$ pp; 95\% CI $\RcvDCI$ pp} \\
\bottomrule
\end{tabular}
\caption{\textbf{Coverage, repeatability, and selection gain separate.}
MATH-500, selected panel versus the same-code control
(\S\ref{sec:math-setup}). $\widehat H$ uses repeat means; $G$ tests
choices on held-out repeats. $D$ is their paired, unrounded
between-arm difference, so displayed means may not subtract exactly.
}
\label{tab:clone-control}
\end{table}

\subsection{Repeatable Score Patterns Are Dominated by Losses}
\label{sec:repeat-result}
Generated programs have more repeatable scored-outcome patterns than
clones. Their mean residual correlation is $\RcvRealRho$, compared
with $\RcvCloneRho$ for clones; the paired-bootstrap difference is
$\RcvRhoDiff$ (95\% interval $\RcvRhoDiffCI$).
The correlation measures repeatability after removing additive member
and task effects (\S\ref{sec:protocol}).

Persistent differences overwhelmingly favor \bare{}.
All-three-repeat losses span $\RcvLossTasks$ tasks
($\RcvLossPairs$ member--task pairs), whereas all-three-repeat wins
occur on one task shared by all eight generated members.
That apparent repair is sensitive to answer-format extraction
(Appendix~\ref{app:repeatability}).
This directional asymmetry explains how strong score repeatability
can coexist with few observed wins.

The winning task illustrates why the scoring path matters. Its gold
answer is \texttt{12}. The baseline ends with
\texttt{\#\#\#\# 12th grade}, whose complete final-answer payload is
retained and scored incorrect. A generated member returns
\texttt{12th grade}, which instead reaches numeric fallback and is
scored correct. The recorded advantage therefore reflects an
answer-format difference. We retain the original scores; this example
explains the sole persistent winning task, without assigning a common
cause to the persistent losses.

Held-out selection provides no clear positive evidence beyond clones.
The clone-adjusted mean is $D=\RcvD$ pp (Table~\ref{tab:clone-control}),
with randomization $p$-value $p=\RcvP$. Its interval includes zero and the positive-support
conditions are not met, giving \textsc{Insufficient Evidence}
rather than an equivalence conclusion.
Excluding the three tasks containing conflicting scores gives
$D=\RcvConflictD$ pp on $\RcvConflictN$ paired tasks and the same
decision (Appendix~\ref{app:wp1r}).

\paragraph{Three repeats have limited power for crossovers.}
\label{sec:power}
In the two-specialist setting described in \S\ref{sec:protocol},
the simulated diagnostic has $0\%$ power at $R=3$ and $100\%$
at $R=10,20$, despite true $H_{\rm stable}=17.5$ pp.
These are observed detection rates over $300$ simulated datasets at
$R=3$ and $150$ at each higher repeat count. Smaller cyclic
crossovers remain largely undetected even at $R=20$
(Appendix Table~\ref{tab:calibration}). Increasing repeats therefore
helps in the strong positive case without ensuring sensitivity to
every crossover. The result characterizes this diagnostic and these
settings, rather than all procedures using three repeats. It explains
why the empirical nonpositive result leaves stable complementarity
unresolved. We next evaluate the tested pre-execution decision rule
directly, whose realized gain is a separate question.

\subsection{The Tested Selector Adds No Gain and Full-Budget Coverage Ties}\label{sec:utility}
\paragraph{The frozen selector matches the development-selected baseline.}
The tested feature policy yields $\RcvPolicyGain$ pp gain. Its frozen
classifier directly selects \bare{} on every evaluation
task; the confidence-based fallback is never triggered. These choices
give $\RcvBaselineAcc\%$ accuracy over three repeats, identical to the
development-selected fixed member.
Appendix~\ref{app:selection} gives its construction and the
same-set comparator results.

\paragraph{At the full execution budget, the oracle coverages tie.}
\label{sec:selection-result}
At $b=27$ executions, the development-ranked portfolio and same-code
controls both reach $\RcvReplayFullCoverage\%$ oracle coverage
(Table~\ref{tab:execution-means}).
At $b=9$ and $15$, the portfolio-minus-clone differences are
$\RcvReplayNineGap$ pp and $\RcvReplayFifteenGap$ pp.
At $b=3$, the portfolio contains only \bare{}, so the
$\RcvReplayThreeGap$ pp difference reflects execution variation
between identical programs.

\begin{table}[htbp]
\centering\small
\begin{tabular}{@{}lrrrr@{}}
\toprule
Harness executions per task $b$ & 3 & 9 & 15 & 27 \\
\midrule
Panel members, including \bare{} & 1 & 3 & 5 & 9 \\
Panel oracle coverage (\%) & 98.45 & 98.70 & 98.70 & 98.70 \\
Clone oracle coverage (\%) & 98.24 & 98.63 & 98.69 & 98.70 \\
Panel minus clones (pp) & $+0.20$ & $+0.08$ & $+0.02$ & $+0.00$ \\
\bottomrule
\end{tabular}

{\scriptsize Differences are computed before rounding the displayed coverage values.}

\caption{\textbf{Full-budget oracle coverage ties in the recorded executions.}
MATH-500 coverage from the development-ranked panel and uniform
same-code sampling. Clone values are exact subset means
(\S\ref{sec:selection-design}). Budgets count harness executions;
model-call and token costs vary by program.}
\label{tab:execution-means}
\end{table}

Together, the selector and replay results separate two operational
questions: the tested pre-execution policy yields zero gain, while
post-execution oracle coverage ties at the full recorded budget.

\subsection{BIRD Traces Locate Breaks Between Code and Execution}\label{sec:supporting}\label{sec:f1-collapse}
Source variation can disappear before it produces a valid final answer.
The supporting BIRD study follows this process directly in independent
program populations, using its task-specific judging and protocols.
Here a mechanism is an implemented prompt transformation, tool use,
or control-flow operation. This execution view complements the
MATH controls by examining what the programs actually do.

\paragraph{Different source code often produces identical scored outcomes.}
All $12$ discovery candidates pass a code-difference check, yet
$21/66$ generated-candidate pairs have identical correctness vectors
($36/91$ including baselines). Source similarity has essentially no
Spearman rank correlation with correctness disagreement ($\rho=-0.010$).
This agreement on $60$ tasks motivates examining the traces
(Appendix Figure~\ref{fig:phenomenon}).

\paragraph{Traces distinguish absent, inactive, and broken mechanisms.}
\textbf{T1: described but absent.} Six of twelve candidates describe
retrieval or self-checking but make a single model call without an
implemented retrieval or self-checking step. This classification covers all twelve candidates;
the five-harness trace audit supplies worked examples.

\textbf{T2: present but untriggered.} For the execution-repair baseline,
the first SQL executes successfully on all ten audited tasks, so the repair branch
never fires and final SQL equals \bare{}. A successful execution
can therefore leave a semantically wrong answer unchanged.

\textbf{T3: executed but broken.} One candidate runs multi-turn
repair but returns prose as final SQL, losing tasks that the
baseline answers correctly. The three cases locate distinct checks:
implementation, branch activation, and valid output.
Appendix~\ref{app:bird-traces} provides worked traces and hand-written
controls, with their development-exposure limitations.

The separate BIRD admission study tests whether prescribing strategies
and screening programs increases headroom. Its implemented gate combines
structural exclusion and conformance verification. Forced/gated minus
free/ungated generation gives $\RevisionPrimaryDelta$\,pp, with no
increase in observed mean headroom (Appendix~\ref{app:phase2-protocol}).
The admission comparison measures what screening changes; the traces
locate where proposed operations fail to deliver useful answers.
Together, these BIRD studies locate execution failures that complement
the MATH evidence about persistence and selection.

\section{Discussion and Limitations}\label{sec:discussion}
A program population can provide additional answers and highly
repeatable score patterns while leaving useful specialization
unestablished. Our same-code control exposes the first gap:
byte-identical programs have zero stable complementarity by construction,
yet produce positive oracle headroom even after averaging repeats.
This headroom captures a real opportunity to obtain correct answers.
Attributing that opportunity to program identity, however, requires
comparison with what further executions of a fixed program already
provide. The central evaluation question is therefore what changing
programs adds when additional sampling is also available.

The generated panel's strong residual correlation chiefly identifies
recurrent weaknesses in the recorded scores.
Equation~\ref{eq:identity} clarifies why the direction matters:
variation in the size of a disadvantage can be highly repeatable
without changing which member is preferable. Repeatability is
informative about where a population fails; demonstrating a division
of labor additionally requires different members to retain advantages
on different tasks.

\paragraph{Mechanisms should target identifiable failures.}
The separate BIRD traces make the design problem concrete. An operation
must be implemented, activated on the relevant input, and produce a
valid answer before it can contribute to performance. The inactive
repair branch illustrates a mismatch between a mechanism and the
failure it needs to address: SQL can execute successfully while
answering the question incorrectly, leaving an execution-error trigger
unable to address that failure. Designing a mechanism therefore also
means specifying the task conditions under which it should change an
answer. A targeted follow-up would compare enabled and disabled
versions of the same mechanism on tasks that exercise its intended
trigger, then test any advantage on fresh executions. Such
interventions would connect intended behavior to a causal contribution
and test whether that contribution persists.

\paragraph{Population value depends on a usable decision.}
For pre-execution routing, useful task advantages must be recognizable
from information available before the answer is known. A portfolio
that runs several programs instead needs an answer-selection or
aggregation procedure that can turn its outputs into a better final
answer. Our frozen selector and oracle replay address different parts
of this problem: the former retains \bare{} on every task, while the
latter measures coverage after correctness is known. Their respective
zero gain and full-budget coverage tie make additional fixed-program
executions an essential comparator. Future evaluations should combine
this comparator with repeat counts calibrated to detect task-wise
crossovers and an implementable decision rule under a matched resource
budget. This would establish whether the advantages associated with
program identity improve inference beyond the sampling opportunities
a larger population creates.

\paragraph{Limitations and unresolved questions.}
The MATH evidence concerns one solver, a selected low-call panel,
and three repeats. Known-truth calibration shows that this repeat
count can miss substantial task-wise crossovers for the tested diagnostic. The
\textsc{Insufficient Evidence} decision leaves stable complementarity
unresolved and does not establish equivalence.
The repeat diagnostic and common-complete-set analysis were developed
after collection; recovery retained the programs, tasks, solver, and
repeat count. The $\RcvExcludedN$ excluded tasks are harder than the
$\RcvPairedN$ included tasks: baseline repeat-mean accuracy is $71.43\%$
on excluded tasks and $95.85\%$ on included tasks
(Appendix~\ref{app:wp1r}). Recorded scores include answer-format
effects. The selector result concerns the
tested policy and features. Replay matches harness executions, with
model calls and tokens varying by program, so the observed tie does
not establish equal-cost utility. BIRD uses separate populations and
scoring conditions; its traces identify execution failures without
establishing the causes of the MATH losses.

\section{Conclusion}\label{sec:conclusion}
The value of a harness population depends on how its differences improve
inference. Our same-code control and repeated executions show why
coverage, repeatability, and usable specialization require separate
evidence: identical programs yield positive oracle headroom, while
generated programs exhibit persistent differences dominated by losses.
On the tested MATH panel, the frozen selector gains $\RcvPolicyGain$ pp,
and repeated baseline executions match the portfolio's
$\RcvReplayFullCoverage\%$ oracle coverage at $27$ harness executions.
Stable complementarity remains unresolved at three repeats.

These findings give automated harness design a concrete objective:
develop reproducible task advantages that a usable decision rule can
exploit. To justify additional programs through specialization,
evaluation must show what those decisions gain over spending the
relevant inference budget on further executions of a fixed program.
\label{endofmain}

\subsection*{Ethics Statement}
All experiments use public BIRD and MATH-500 benchmark data and generated
program artifacts; no human participants, private records, or personally
identifiable information are collected.  Model calls are reported as part of
the experimental provenance, and the paper does not release credentials or
provider responses that would expose private data. The results distinguish
recorded answer coverage from the performance of a selector operating
before execution.

\subsection*{Reproducibility Statement}
The manuscript specifies the populations, task splits, execution
conditions, scoring rules, statistical procedures, and sensitivity
analyses used for the reported results. The appendices provide
additional protocol details, calibration results, and worked examples.

\subsection*{AI use statement}
Generative AI tools assisted with research questions, experimental and
statistical design, counterexample analysis, code development, model-API
workflows, data analysis, figures, and manuscript preparation. Versioned
artifacts record the analyses and revisions; automated tests, local runtime
checks, and same-model subagents supported mechanical verification.
The human authors set the research direction and retain responsibility
for scientific decisions, verification, the final content, and submission.

\bibliography{references}
\bibliographystyle{iclr2027_conference}

\appendix
\raggedbottom
\FloatBarrier
\needspace{12\baselineskip}
\section*{Appendix Guide}
The appendices document program generation, repeated execution, selection,
calibration, and the supporting BIRD studies.
\begin{description}
\item[\ref{app:notation}] \hyperref[app:notation]{Notation summary}.
\item[\ref{app:inventory}] \hyperref[app:inventory]{Harness generation, program inventory, and data splits}.
\item[\ref{app:wp1r}] \hyperref[app:wp1r]{Same-code controls and repeated-execution protocol}.
\item[\ref{app:statistics}] \hyperref[app:statistics]{Statistical analysis, calibration, and sensitivity}.
\item[\ref{app:selection}] \hyperref[app:selection]{Frozen selector and execution-matched replay}.
\item[\ref{app:bird-traces}] \hyperref[app:bird-traces]{BIRD execution traces and supporting analyses}.
\item[\ref{app:phase2-protocol}] \hyperref[app:phase2-protocol]{BIRD admission-policy study}.
\end{description}
\FloatBarrier
\section{Notation Summary}\label{app:notation}
Table~\ref{tab:notation} collects the notation used in the main text and
the analysis appendices. Mathematical scores and gains use proportions;
reported accuracies use \% and reported gains use percentage points (pp).
Thus a gain of $0.01$ is $1$\,pp. Population expectations, finite-sample
estimates, and held-out gains are distinct quantities.

\begingroup
\small
\setlength{\tabcolsep}{5pt}
\renewcommand{\arraystretch}{1.12}
\begin{longtable}{@{}>{\raggedright\arraybackslash}p{0.24\linewidth}>{\raggedright\arraybackslash}p{0.72\linewidth}@{}}
\caption{Symbols, definitions, and scope.}\label{tab:notation}\\
\toprule
Notation & Meaning \\
\midrule
\endfirsthead
\multicolumn{2}{l}{\textbf{Table~\thetable{} (continued)}}\\
\toprule
Notation & Meaning \\
\midrule
\endhead
\midrule
\multicolumn{2}{r}{Continued on next page}\\
\endfoot
\bottomrule
\endlastfoot
\multicolumn{2}{@{}l}{\textbf{Members, tasks, and outcomes (\S\ref{sec:measures})}}\\*
$P$, $h$, $M$ & Harness population, an indexed member $h\in P$, and population size $M=|P|$. Identical clone slots have distinct member identities. \\
$H_0$, $H_1$ & Single-call baseline and the other member in the two-member identity. \\
$x$, $x_i$, $i$, $n$ & A task, evaluated task $i$, its index, and the number of evaluated tasks. \\
$r$, $R$ & Execution-repeat index and number of repeats per member--task pair; empirical repeat comparisons use $R=3$. \\
$Y(h,x,r)$ & Binary correctness of member $h$ on task $x$ in repeat $r$. \\
$p_h(x)$; $p_0(x),p_1(x)$ & Success probability over fresh executions at fixed conditions. For two members, $p_j(x)=p_{H_j}(x)$, $j\in\{0,1\}$. \\
$\hat p_h(x)$ & Complete-repeat estimate $R^{-1}\sum_{r=1}^R Y(h,x,r)$. \\
$\mathbb E_x$, $\Pr$ & Expectation over the target task distribution and probability. In the Jensen argument, unsubscripted $\mathbb E$ instead averages execution sampling at a fixed task. \\
$H(P)$ & Single-run oracle accuracy minus the best fixed member's accuracy on the same tasks and repeat (Equation~\ref{eq:headroom}). \\
$\widehat H$ & Repeat-averaged plug-in headroom: replace $Y$ in $H(P)$ by $\hat p_h(x)$ before taking either maximum. \\
$H_{\rm stable}$ & Population headroom computed from the success probabilities $p_h(x)$. \\
$\Delta(x)$ & Two-member success-probability difference $p_1(x)-p_0(x)$ (Equation~\ref{eq:identity}). \\
\addlinespace
\multicolumn{2}{@{}l}{\textbf{Repeat diagnostics, selection, and budgets}}\\*
$G$; $G_{\rm real},G_{\rm clone}$ & Mean held-out accuracy gain of the discovery-selected task-wise mapping over the discovery-selected fixed member; subscripts identify the generated-plus-baseline and same-code arms (\S\ref{sec:protocol}). \\
$D$ & Paired clone-adjusted gain $G_{\rm real}-G_{\rm clone}$, a finite-repeat diagnostic (Equation~\ref{eq:clone-difference}). \\
$\rho_{\rm res}$ & Mean Pearson correlation of residual score patterns between each held-out repeat and the other repeats' mean (Appendix~\ref{app:repeatability}); labeled residual $r$ in Figure~\ref{fig:sampling-control}. \\
$\rho$ & BIRD Spearman rank correlation between source similarity and correctness disagreement (\S\ref{sec:supporting}). \\
$p$; CI & Randomization $p$-value for the MATH diagnostic; confidence interval. A scalar $p$-value is distinct from success probability $p_h(x)$. \\
$Z(x)$, $\pi_Z(x)$ & Pre-execution task features and the member selected from them; the implemented features use question text and training-fold member accuracies. \\
$h_{\rm dev\text{-}fixed}$ & Fixed member chosen by development accuracy, before evaluation. \\
$S(\pi_Z)$ & Evaluation-task mean of repeat-mean accuracy differences between the frozen feature selector and $h_{\rm dev\text{-}fixed}$ (Appendix~\ref{app:selection}). \\
$b$ & Number of harness executions per task in oracle replay, $b\in\{3,9,15,27\}$; distinct from model-call and token counts. \\
$k$ in Pass@$k$ & Number of sampled answers; Pass@$k$ measures the probability that at least one is correct. \\
\addlinespace
\multicolumn{2}{@{}l}{\textbf{Additional notation in the analysis appendices}}\\*
$h_0$, $A_r(h,x)$ & Arm-specific reference (single-call or clone-c1) and baseline-relative score $Y(h,x,r)-Y(h_0,x,r)$ (Appendix~\ref{app:repeatability}). \\
$T$ & Number of simulated tasks in calibration; $M$ and $R$ retain their member-count and repeat-count meanings (Appendix~\ref{app:statistics}). \\
$U(0,10^{-9})$ & Uniform distribution used for independent random tie-breaking jitter. \\
$\theta^*$ & A given selector's population gain over the population-best fixed member; bounded above by $H_{\rm stable}$. \\
$\mathcal R_{hx}$, $\bar Y_h(x)$ & Observed development-repeat indices and their mean score. In evaluation, $\bar Y_h(x)$ averages all three repeats (Appendix~\ref{app:selection}). \\
$y(x)$ & Member-ID training label maximizing $\bar Y_h(x)$; distinct from binary correctness $Y(h,x,r)$. \\
$C$ & Inverse regularization strength of the logistic classifier, fixed at $1$. \\
$K$, $k$, $q_k(h\mid x)$ & Number of fitted fold models ($1\le K\le5$), model index, and model $k$'s predicted member-class probability. Here $k$ indexes models, rather than answer samples. \\
$\bar q(h\mid x)$; $h_{(1)},h_{(2)}$ & Mean of fitted models' class probabilities; the two highest-probability members. A gap below $0.15$ triggers baseline fallback. \\
$f$, $\binom{f}{b}$ & Incorrect clone executions among $27$ for a task, and the number of size-$b$ subsets containing only incorrect executions; zero when $b>f$. \\
$K_{\rm cand}$, $K_{\rm total}$ & Admitted BIRD candidate count without and with the baseline: $K_{\rm total}=K_{\rm cand}+1$. K-matching matches candidate counts (Appendix~\ref{app:phase2-protocol}). \\
II-A--II-D; A--D & BIRD free/ungated, free/gated, forced/ungated, and forced/gated arms. A--D denote arm headrooms in factorial formulas; their D$-$A contrast differs from the MATH statistic $D$. \\
\end{longtable}
\endgroup

\FloatBarrier
\section{Harness Generation, Program Inventory, and Data Splits}\label{app:inventory}
This appendix describes the MATH population, splits, frozen panel, and model
versions. BIRD protocols appear in Appendices~\ref{app:bird-traces}
and~\ref{app:phase2-protocol}.

\paragraph{Execution interface.}\label{app:execution-interface}
Harnesses implement \texttt{solve(question)} and access the frozen solver
through \texttt{self.llm}. BIRD additionally provides the database schema
as \texttt{self.schema}; \texttt{self.execute(sql)} returns query results
or execution errors. Gold answers remain external to these interfaces.
The execution-repair baseline is labeled \react{} in archived programs
and traces.

\subsection{MATH population, common analysis set, and coverage}\label{app:crossdomain}
The single-run study supplies coverage measurements and programs for the
repeat panel, applying BIRD's free, ungated generation procedure to
competition mathematics with a domain-specific comparator.

\paragraph{Protocol and analysis set.}
\label{app:builder-instruction}
The builder instruction, summarized rather than quoted, requests a Python
\texttt{solve(question)} harness that improves on one greedy call through
\texttt{self.llm}, with freely chosen prompting, decoding, sampling, and
control flow and a final \texttt{\#\#\#\#} answer. Gold answers remain
external. Slots receive no assigned mechanism or diversity target.
Attempts reuse the instruction without prior code or performance
feedback; admission requires a valid interface and nonempty output.

Six builders each receive eight slots under one seed, with at most three
attempts per slot. Parsing, import, interface, and nonempty answers are
checked on two development questions. The target is GLM-5.3-Flash, with
member-specific decoding and temperature $0$ for \bare{}. The
pre-generation split reserves $100$ development problems. Single-run,
repeat, and policy analyses share $\RcvPairedN$ fully observed tasks;
the post-collection completeness rule excludes $\RcvExcludedN$ problems
(Appendix~\ref{app:wp1r}). Development tasks and the selector stay fixed.

The judge compares parsed numbers, including fractions and decimals,
or otherwise normalized LaTeX. Normalization removes units, base
subscripts, and degree symbols, retains text-command content, and maps
matrices to tuples. Gold answers prefixed with \texttt{\#\#\#\#} pass
$\RcvPairedN/\RcvPairedN$ self-comparisons. Single-run baseline accuracy
on this set is $\RevisionMathBest\%$.

The gate admits $35/48$ slots. GLM-5.3 and Qwen3.8-Max exhaust the
$32$k-token generation budget before emitting code in most slots,
leaving their styles under-represented in this fixed-budget population.

\paragraph{Results on the common complete set.}
All $36$ members, including \bare{}, have one score per task:
$\RcvSingleCells/\RcvSingleCells$ cells over $\RcvPairedN$ tasks.
Mean pairwise disagreement is $\RevisionMathDisagreement\%$
(maximum $\MathMaxDisagreement\%$), with $\MathUnique/36$ unique
outcome vectors ($\MathGenUnique/35$ generated) and
$\MathIdenticalPairs\%$ identical pairs. The generated union repairs
$\MathUnionRepair\%$ of baseline errors; all generated members fail on
$\MathUnionFail\%$ of baseline-correct tasks. Oracle coverage of
$\RevisionMathOracle\%$ exceeds the best fixed member \bare{}
($\RevisionMathBest\%$) by $\RevisionMathHeadroom$\,pp.
Generated accuracies span $\MathGenMin\%$--$\MathGenMax\%$.

The repeat panel compares eight generated members plus \bare{} with
nine same-code slots on identical tasks. Its held-out diagnostic concerns
this panel; single-run headroom concerns the full population. Both the
archived judge and the versioned endpoint-sensitive judge reproduce all
$\RcvSingleCells$ included scores.

\paragraph{Scope.}
These measurements describe one seed and one target on the
completeness-selected MATH set; BIRD uses the separate protocols below.

\FloatBarrier
\subsection{Selected MATH Program Inventory}
\label{app:panel-mechanisms}
Table~\ref{tab:panel-mechanisms} describes the nine frozen members
selected for the repeat study. H1--H8 are neutral labels for generated
members, all constructed under seed $0$. The listed operations describe
what each program implements, not whether every branch fires or improves
correctness. The mean-call eligibility ceiling is not a per-execution
maximum, and one logical call may request several samples.

\begin{table}[htbp]
\centering\small
\begin{tabular}{@{}p{0.29\linewidth}p{0.66\linewidth}@{}}
\toprule
Frozen member & Source-defined execution path \\
\midrule
\bare{} & One temperature-$0$ solution with the final-answer format. \\
H1 (DeepSeek) & Three temperature-$0.4$ solutions; normalized-answer vote; a greedy fallback when no answer or a tie remains. \\
H2 (ERNIE) & Draft a solution, then ask the solver to verify/correct the extracted answer; return the checked answer. \\
H3 (GLM) & Up to five differently prompted temperature-$0$ solutions; stop when three agree; adjudicate unresolved disagreement, with format fallbacks. \\
H4 (Kimi) & Request five temperature-$0.7$ samples in one call; extract answers and vote. \\
H5 (Kimi) & Initial solution, verification, and an alternative solution; vote, with an additional resolution call if no majority emerges. \\
H6 (MiniMax) & Draft, critique, and an additional tie-breaking call on disagreement; parse failures use an earlier available answer. \\
H7 (MiniMax) & Request five temperature-$0.7$ samples in one call; normalize and vote, resolving ties by first occurrence. \\
H8 (Qwen) & Draft plus an independent solve-and-audit call; compare answers and request adjudication on conflict, with a format-repair path. \\
\bottomrule
\end{tabular}
\caption{\textbf{What was selected for the repeat study.} All calls use
the same frozen solver; verification and adjudication are solver prompts,
not access to benchmark labels. Six seats are builder-stratified; H5 (Kimi) and H7 (MiniMax)
are the two at-large seats.}
\label{tab:panel-mechanisms}
\end{table}

\FloatBarrier
\subsection{Models and versions}\label{app:models}

Table~\ref{tab:models} gives the exact API model identifiers. All roles
use one commercial OpenAI-compatible aggregator. Target identity is fixed;
decoding follows each harness, with temperature $0$ for the single-call
baseline. Phase-II generation uses temperature 0.7 and a 16{,}384-token
budget; Phase-I conditions appear in Appendix~\ref{app:phase1-provenance}.
Phase I uses a proxy execution judge, Phase II the official BIRD scorer
(30\,s timeout) with a secondary legacy-judge audit, and MATH-500 the
answer comparator in Appendix~\ref{app:crossdomain}. Reported accuracy
uses no LLM judge.

\begin{table}[H]
\centering
\small
\begin{tabular}{p{0.26\linewidth}p{0.50\linewidth}p{0.14\linewidth}}
\toprule
Role & Model (exact version string) & Used in \\
\midrule
Frozen target solver & \texttt{GLM-5.3-Flash} (baseline temperature $0$) & MATH / BIRD \\
BIRD discovery builder & \texttt{GLM-5.3-Flash} & App.~\ref{app:phase1-provenance} \\
MATH / BIRD admission builders & \texttt{GLM-5.3}, \texttt{Qwen3.8-Max},
\texttt{DeepSeek-V4-Pro}, \texttt{Kimi-K3}, \texttt{MiniMax-M3},
\texttt{ERNIE-5.0-Thinking-Preview} & MATH / BIRD \\
Judge & proxy / official BIRD execution scorer; MATH answer comparator & see text \\
\bottomrule
\end{tabular}
\caption{Model inventory for the reported benchmark studies. The target is held frozen within each
phase; Phase-II builders come from six independent labs.}
\label{tab:models}
\end{table}

\FloatBarrier
\section{Same-Code Controls and Repeated-Execution Protocol}\label{app:wp1r}
All MATH primary estimates use the same $\RcvPairedN$ fully observed tasks.
The protocol below specifies the panel, inclusion rule, and repeat analysis.

\paragraph{Panel and inclusion.}
Programs, solver, judge, development split, and three repeat identities
are fixed. A salted hash selects one low-call member per builder plus
two at-large members and \bare{}; the control contains nine identical
baseline slots. The September 26, 2026, 05:15 UTC snapshot combines
generated continuation 10 and clone continuation 20. After outcome
inspection, we retained tasks with a recorded binary score in every
required single-run and repeat cell. These $\RcvPairedN$ task identities
define the common reanalysis set.

\paragraph{Completeness and exclusions.}
The single-run matrix contains $36\times\RcvPairedN=\RcvSingleCells$
observed cells and each repeat arm $9\times3\times\RcvPairedN=\RcvCells$,
giving $100\%$ completeness for all three matrices. Policy evaluation and
oracle replay use the same task identities.

The $\RcvExcludedN$ excluded problems contain
$\RcvOriginalMissingCells$ unscored generated cells, distributed as
DeepSeek 5, ERNIE 4, GLM 9, Kimi 30, MiniMax 9, and Qwen 4; baseline
and clone records are complete. Their pre-subsetting fraction,
$\RcvOriginalExcludedPct\%$, is below the frozen $10\%$ gate.
Excluded tasks are harder and longer, with lower baseline accuracy
(Table~\ref{tab:complete-subset}). Unknown scores and all incurred
acquisition costs remain in the source records.

\begin{table}[H]
\centering\small
\begin{tabular}{@{}lrr@{}}
\toprule
Observed characteristic & Included ($\RcvPairedN$) & Excluded ($\RcvExcludedN$) \\
\midrule
Baseline repeat-mean accuracy (\%) & 95.85 & 71.43 \\
Mean benchmark difficulty level & 3.47 & 4.29 \\
Level-5 tasks (\%) & 26.42 & 64.29 \\
Mean question length (characters) & 198.75 & 284.71 \\
\bottomrule
\end{tabular}
\caption{\textbf{Observed characteristics of included and excluded tasks.}
Excluded tasks have higher benchmark difficulty, longer questions, and
lower baseline accuracy.}
\label{tab:complete-subset}
\end{table}

\paragraph{Provenance and conflicts.}
Manifests are checked on 13 fields covering execution conditions,
programs, task/split hashes, panel draw, and schedule; snapshots,
matrices, and task lists are hash-bound. Completed scores replace
unknowns, with first-completed precedence on disagreement. Five
generated cell identities conflict, including three on three retained
tasks; clones have none. Collector and concurrency changes are logged.

\paragraph{Estimates and sensitivity.}
Recorded binary scores give plug-in headroom of $\RcvRealPlugin$\,pp
(generated) and $\RcvClonePlugin$\,pp (clone), and
$D=\RcvD$\,pp (95\% interval $\RcvDCI$\,pp;
\textsc{Insufficient Evidence}). Removing the three conflict-affected
tasks from both arms and reselecting on the remaining $\RcvConflictN$
tasks gives $D=\RcvConflictD$\,pp (interval $\RcvConflictDCI$\,pp),
with the same decision.

\subsection{Exploratory Repeatability and Score Sensitivity}\label{app:repeatability}
For non-reference members, define $A_r(h,x)=Y(h,x,r)-Y(h_0,x,r)$,
where $h_0$ is \bare{} or clone-c1. Within each repeat, subtract row
and column means and add the grand mean. Correlate each held-out
residual matrix with the residual of the other two repeats' mean,
then average the three Pearson correlations to obtain $\rho_{\rm res}$. A paired task bootstrap uses
$2{,}000$ shared draws. Correlations are $\RcvRealRho$
($\RcvRealRhoCI$) for generated members and $\RcvCloneRho$
($\RcvCloneRhoCI$) for clones, with difference $\RcvRhoDiff$
($\RcvRhoDiffCI$). This post-collection diagnostic measures repeatable
score patterns after removing additive effects.

Losses in all three repeats span $\RcvLossPairs$ member--task pairs on
$\RcvLossTasks$ tasks. Wins in all three comprise eight pairs on one
task, shared by all eight generated programs. That task is
format-sensitive: the gold answer is \texttt{12}, baseline outputs end
with \texttt{\#\#\#\# 12th grade}, and a generated member returns
\texttt{12th grade}. The extractor retains the former's complete
\texttt{\#\#\#\#} payload and scores it zero, but uses numeric fallback
for the latter and scores it one. Recorded scores are retained; the
repeated repair here arises from answer formatting.
\FloatBarrier

\FloatBarrier
\section{Statistical Analysis, Calibration, and Sensitivity}\label{app:statistics}\label{app:diagnostics}
Empirical diagnostics use $\RcvPairedN$ complete MATH tasks.
Synthetic calibration uses the sample sizes specified below.

\subsection{Held-out-repeat statistic and decision rule}
In each of three folds, two discovery repeats select a member for each
task and a single best fixed member by mean discovery accuracy. The
third repeat scores the task-wise mapping against that fixed member.
Averaging task-level gains over folds gives $G$; the paired difference
between arms is $D=G_{\rm real}-G_{\rm clone}$
(\S\ref{sec:calibration-test}). The repeat-averaged oracle plug-in
$\widehat H$ is descriptive and does not determine the decision.

Discovery scores lie on $\{0,\tfrac12,1\}$, so ties are common.
Selection uses independent, identically distributed uniform jitter $U(0,10^{-9})$ to break member ties
symmetrically, averaging ten frozen tie realizations for the empirical
report. Each calibration dataset uses one tie realization.
For the empirical randomization reference, member labels are permuted
independently within each task and each repeat in each arm. Both
selections, both gains, and $D$ are recomputed for each of $10{,}000$
draws, with the $+1$ Monte Carlo correction. A paired task bootstrap
uses $2{,}000$ draws with a fixed random seed and the same resampled
task identities in both arms to obtain the 95\% interval.

Positive support requires: at most $10\%$ incomplete tasks in either
arm or their intersection before subsetting; at least $30$ paired
tasks; randomization $p<0.05$; positive bootstrap lower bounds for
$D$ and $G_{\rm real}$; and $D>1.0$\,pp. Repeats must have matched
conditions and verified independent execution provenance.
The current decision is \textsc{Insufficient Evidence}, rather
than an equivalence finding.

\paragraph{Analysis specification.}
The acquisition protocol and initial stopping rule preceded collection;
the v3 diagnostic, recovery, and completeness-selected reanalysis are
post-collection amendments. Estimates use the September 26 snapshot;
inclusion, exclusions, and conflict sensitivity are specified in
Appendix~\ref{app:wp1r}.

\subsection{Population target and limits of the diagnostic}
For member success probabilities $p_h(x)$, stable complementarity is
\[
H_{\rm stable}=\mathbb E_x\max_h p_h(x)-\max_h\mathbb E_x p_h(x).
\]
It is zero when some fixed member is expectation-best on almost every
task. For a given selector, its population gain $\theta^*$ relative to the
\emph{population-best fixed} member satisfies
$\theta^*\le H_{\rm stable}$, because its success probability on each
task is at most $\max_h p_h(x)$.

Empirical $G$ uses a fixed member selected from finite discovery data.
Clone subtraction provides a sampling reference, while $D$ and its
interval remain distinct from $H_{\rm stable}$: a near-zero $D$ does
not bound the population target near zero.

The randomization reference is exact under equal-ability member-label
exchangeability. For the broader global-dominance null, error control
is empirical over the calibration grid.

\paragraph{Finite-repeat oracle estimates.}
For repeat means $\hat p_h(x)$, with $\mathbb E$ here taken over repeated-execution sampling at fixed $x$, Jensen's inequality gives
$\mathbb E\max_h\hat p_h(x)\ge\max_h\mathbb E\hat p_h(x)$
on each task. Headroom subtracts a second maximum, so this inequality
does not imply upward bias of the difference in general. In a
known-truth equal-ability simulation with $p_h(x)=0.9$, $M=9$ members, $T=400$ tasks, and
$R=3$, the mean plug-in is $8.69$\,pp over $300$ datasets despite
$H_{\rm stable}=0$. The empirical same-code arm likewise gives
$\RcvClonePlugin$\,pp. These examples show that positive plug-in
headroom is insufficient evidence of specialization.

\subsection{Known-truth calibration and statistical power}
Binary tensors are sampled from specified member--task probability
matrices. Null settings include equal ability, shared task difficulty,
global dominance, the $0.5/0.9$ dominance configuration, and heterogeneous
dominance. Alternatives
include crossovers over latent task types and difficulty-related
partial crossover. The grid spans $M\in\{3,9,20\}$,
$T\in\{100,400,1000\}$, $R\in\{3,5,10,20\}$, complete data,
5\% missing completely at random (MCAR), and 3\% hardness-related missingness.

For every repeat count $R$, the calibrated statistic uses $R$ folds:
each fold learns both choices from the other $R-1$ repeats and scores
them on its single held-out repeat. The table's \emph{plug} comparator
declares a positive result when the lower endpoint of its percentile
95\% task-bootstrap interval exceeds zero ($1{,}000$ draws).
It fixes the task-wise maxima and sample-best member before resampling.
The simulation variant, labeled v3$^\ast$, uses the positive-support
conditions above with $300$ bootstrap draws and $300$ randomizations,
but checks the completeness gate only on the real arm. The empirical
rule additionally gates the clone arm and paired intersection.
Simulation abstention denotes failure to obtain positive support.

In the two-specialist setting, the $400$ tasks form two equal halves.
Each of two specialists succeeds with probability $0.95$ on its own
half and $0.60$ on the other; the remaining seven members have
probability $0.60$ throughout. Outcomes are independent Bernoulli
draws. The oracle mean is $0.95$ and the best fixed mean is $0.775$,
so $H_{\rm stable}=17.5$\,pp. All nine clone slots share the first
specialist's probability profile, with fresh independent draws.

Table~\ref{tab:calibration} reports $M=9,T=400$, using $300$
Monte Carlo datasets per cell ($150$ in sweeps) and $300$
randomizations per dataset. Across the $100$-cell grid, the largest
observed null false-support rate is $2\%$. A two-specialist
alternative with $H_{\rm stable}=17.5$\,pp has power $0\%$ at
$R=3$, approximately $0.7\%$ at $R=5$, and $100\%$ at $R=10,20$;
smaller cyclic crossovers remain largely undetected at $R=20$.
These setting-specific results show why three repeats leave stable
complementarity unresolved.

\begin{table}[H]
\centering\small
\renewcommand{\arraystretch}{0.95}
\setlength{\tabcolsep}{4pt}
\footnotesize
\begin{tabular}{@{}lccccccc@{}}
\toprule
 & \multicolumn{2}{c}{$R{=}3$} & \multicolumn{2}{c}{$R{=}10$} & \multicolumn{2}{c}{$R{=}20$} & abst. \\
\cmidrule(lr){2-3}\cmidrule(lr){4-5}\cmidrule(lr){6-7}\cmidrule(lr){8-8}
regime & plug & v3$^\ast$ & plug & v3$^\ast$ & plug & v3$^\ast$ & $R{=}3$ \\
\midrule
equal $p{=}0.9$$^\dagger$ & 100 & 0 & 100 & 0 & 100 & 1 & 100 \\
equal + difficulty$^\dagger$ & 100 & 0 & 100 & 1 & 100 & 1 & 100 \\
$0.5/0.9$ dominance$^\dagger$ & 100 & 0 & 100 & 0 & 13 & 0 & 100 \\
dominance$^\dagger$ & 100 & 0 & 100 & 0 & 100 & 0 & 100 \\
cross $H{\approx}0.9$\,pp & 100 & 1 & 100 & 0 & 100 & 0 & 99 \\
cross $H{\approx}1.8$\,pp & 100 & 1 & 100 & 1 & 100 & 0 & 99 \\
cross $H{\approx}4.2$\,pp & 100 & 0 & 100 & 0 & 100 & 0 & 100 \\
two strong specialists, $H{=}17.5$\,pp & 100 & 0 & 100 & 100 & 100 & 100 & 100 \\
\bottomrule
\end{tabular}

{\scriptsize $^\dagger$Null ($H{=}0$): false-support \%; otherwise power \%. $M{=}9,T{=}400$, $300$ datasets at $R{=}3$ ($150$ in sweeps). \emph{plug}: positive 95\% bootstrap lower bound. $^\ast$v3 simulation variant: real-arm completeness gate, $300$ bootstrap draws and $300$ randomizations. The empirical rule additionally gates clones and their paired intersection; details are given above.}

\caption{\textbf{Three repeats can miss even a strong positive case.}
At $M=9,T=400$, null rows report false-support \% and alternative
rows report power \%; the final column gives $R=3$ abstention.
The simulated clone-calibrated rule (v3$^\ast$) rarely declares positive
support when stable complementarity is zero, but has little sensitivity
to the tested small and moderate crossovers.}
\label{tab:calibration}
\end{table}

\FloatBarrier
\section{Frozen Selector and Execution-Matched Replay}\label{app:selection}
This appendix specifies how development data determine the frozen
program choices and portfolio ranking, and how each is evaluated.

\subsection{Frozen feature selector}
The reported selector $\pi_Z$ uses multinomial logistic regression with
five task folds on the $100$ development tasks. The score summary for
member $h$ on development task $x$ is the available-repeat mean,
\[
 \bar Y_h(x)=\frac{1}{|\mathcal R_{hx}|}
             \sum_{r\in\mathcal R_{hx}}Y(h,x,r),
\]
where $\mathcal R_{hx}$ contains its observed development repeats.
A task is eligible for fitting only when every member has at least one
observed repeat. Its training label is
$y(x)=\arg\max_h\bar Y_h(x)$, with ties resolved by ascending member ID.
Eligible tasks, ordered by task ID, are split into five consecutive
folds without shuffling; all repeats of a task stay together.

The TF-IDF vocabulary and inverse-document frequencies are fitted once
on eligible development questions, using English stop-word removal and
at most $300$ features. Each fold model appends the vector of member
accuracies computed on its four training folds to every question's
features. This vector is constant across questions within that model;
only the text features vary by task. Logistic regression uses inverse regularization strength $C=1$ and at most $1{,}000$ iterations. A training fold with only one
label is omitted; absent member classes receive zero probability.
If $1\le K\le5$ models are fitted and $q_k(h\mid x)$ is model $k$'s
predicted class probability, inference uses
$\bar q(h\mid x)=K^{-1}\sum_{k=1}^Kq_k(h\mid x)$.
Let $h_{(1)}$ and $h_{(2)}$ have the two highest averaged probabilities.
The final frozen choice is
\[
 \pi_Z(x)=
 \begin{cases}
  \text{\bare{}}, & \bar q(h_{(1)}\mid x)-\bar q(h_{(2)}\mid x)<0.15,\\
  h_{(1)}, & \text{otherwise}.
 \end{cases}
\]
The threshold is applied after averaging; a tied top probability
therefore triggers the baseline fallback. The policy
table evaluates these final choices, including the baseline fallback.
It reports repeat-mean accuracy, with harness executions as the unit of
evaluation; a selected harness can make multiple model calls.

Comparators are \bare{}, the development-selected fixed member,
one fixed member drawn once with seed $20260915$
(\texttt{gsm\_kimi\_s0\_g3}), and the development-selected member with
baseline fallback on missing cells. All policies are evaluated on the
common $\RcvPairedN$ tasks, with three scores per selected member--task
pair. Paired differences against dev-fixed use $2{,}000$ task-bootstrap
draws. The feature, dev-fixed, and fallback policies retain their
recorded \bare{} choices.

\begin{table}[H]
\centering\small
\begin{tabular}{@{}lcc@{}}
\toprule
Policy & Eval acc.\ (\%) & $S$ vs dev-fixed (pp) [95\% CI] \\
\midrule
\bare{} (fixed) & 95.85 & 0.00 [0.00, 0.00] \\
dev-fixed & 95.85 & --- \\
One randomly drawn fixed member & 86.01 & -9.84 [-13.04, -6.91] \\
Dev-fixed + baseline fallback & 95.85 & 0.00 [0.00, 0.00] \\
$\pi_Z$ (frozen selector) & 95.85 & 0.00 [0.00, 0.00] \\
\bottomrule
\end{tabular}

\caption{\textbf{Frozen selection on MATH-500.} Accuracy is repeat-mean
correctness on fresh executions; $S$ is the paired gain over the
development-selected fixed member, with 95\% task-bootstrap intervals.}
\label{tab:wp1r-policy}
\end{table}

\FloatBarrier
\subsection{Selection gain and the scope of budget utility}
For the implemented selector, $Z(x)$ consists only of question features
and training-fold member accuracies. Its gain over the
development-selected fixed member is
\[
 S(\pi_Z)=\frac1n\sum_{i=1}^n\left[
 \bar Y_{\pi_Z(x_i)}(x_i)-\bar Y_{h_{\rm dev\text{-}fixed}}(x_i)\right],
\]
where $\bar Y_h(x)$ now averages the three evaluation repeats, $n$ is the number of evaluation tasks, and $h_{\rm dev\text{-}fixed}$ is the fixed member selected using development data. We use
$S$ for policy gain and reserve $D$ for the real-minus-clone repeat
diagnostic. Mechanism labels are not inputs to this feature selector.

Replay budget $b$ counts \emph{harness executions}, each potentially
containing multiple model calls; coverage uses a post-execution oracle.

\subsection{Execution-matched replay and collection accounting}
Each arm contains nine members/slots with three repeats. Replay uses the
common $\RcvPairedN$ tasks and frozen development ranking
(\S\ref{sec:selection-result}), taking all three repeats of the first
$b/3$ members for $b\in\{3,9,15,27\}$. With $f$ incorrect clone
executions among $27$, expected coverage of a uniform size-$b$ subset is
$1-\binom{f}{b}/\binom{27}{b}$, where $\binom{f}{b}=0$ for $b>f$.
Table~\ref{tab:execution-means} compares this exact expectation with
generated-panel coverage at matched execution counts.

Acquisition ledgers retain all attempts and known usage, including
excluded tasks and recovery; outcome tables use the common complete set.
Calls and tokens follow each program and acquisition record.

\FloatBarrier
\section{BIRD Execution Traces and Supporting Analyses}\label{app:bird-traces}
Phase I studies BIRD discovery; Phase II studies admission.
This appendix details the discovery programs and their execution paths.
\subsection{Discovery populations and conditions}\label{app:phase1-provenance}

\paragraph{Population and judging.}
Discovery evaluates twelve TTHE-proposer candidates from GLM-5.3-Flash,
plus \bare{} and \react{}, on $60$ tasks (\S\ref{sec:f1-collapse}).
Admission checks import and smoke execution. The baseline flips on
three of $60$ tasks between discovery runs. The set-of-rows proxy judge
agrees with the official BIRD scorer on 0.97--0.99 of outcomes in two
matrices with retained SQL, with no proxy-correct/official-incorrect cases.

\paragraph{Development exposure.}
Hand-written controls were designed after trace inspection, and
$24/151$ evaluation items overlap protocol development. Their headroom
is descriptive. Equal generation budgets cannot be reconstructed from
the early records, which lack generation-tool and per-attempt logs.

\begin{figure}[H]
\centering
\includegraphics[width=0.75\linewidth]{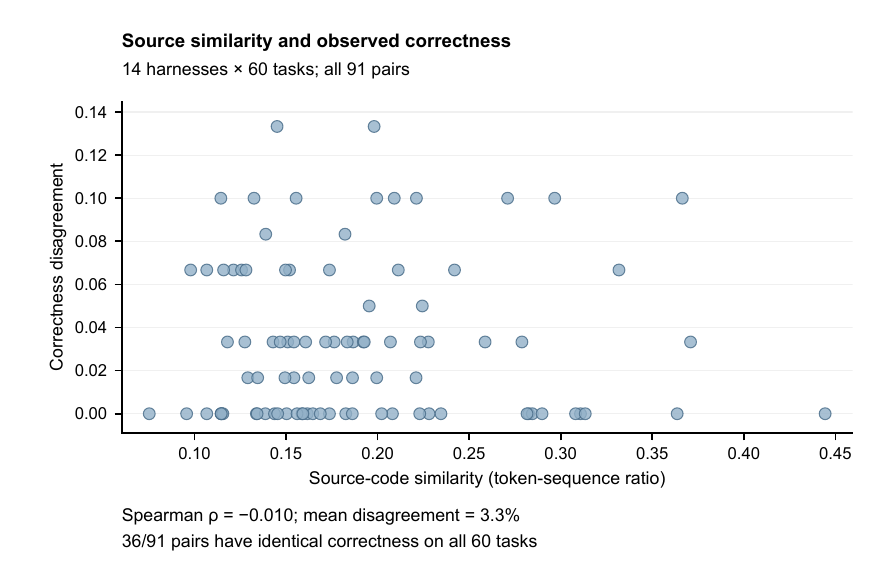}
\caption{\textbf{Source differences need not change observed correctness.}
BIRD discovery: $14$ harnesses, $60$ tasks, one execution each.
Across $91$ pairs including baselines, token-sequence source similarity
has $\rho=-0.010$ rank association with correctness disagreement;
$36/91$ pairs have identical correctness vectors. This is an outcome
comparison on the observed tasks, not a test of program equivalence.}
\label{fig:phenomenon}
\end{figure}
\FloatBarrier
\subsection{Discovery measurements and hand-written controls}\label{app:bird-discovery}

All candidates pass the code-difference check, but $21/66$ generated
pairs share correctness vectors ($36/91$ with baselines).
Source similarity and correctness disagreement have
$\rho=-0.010$ across $91$ pairs (Figure~\ref{fig:phenomenon}).
On these $60$ tasks, mean disagreement is $3.3\%$ overall and
$3.8\%$ among generated candidates, union repair is $7.7\%$,
and single-run headroom is $3.3$\,pp. Traces explain how distinct
programs produce these closely aligned outcomes.

\subsection{Traces distinguish absence, non-activation, and failure}
\textbf{T1: described but absent.} Six of twelve candidates claim
retrieval or self-checking but make one model call without implementing
either step.
\textbf{T2: present but untriggered.} On all ten audited \react{} tasks,
the first SQL executes without error, leaving the repair branch idle and
final SQL identical to \bare{}; execution success can coexist with
semantic error.
\textbf{T3: executed but broken.} One of twelve candidates executes
multi-turn repair but returns prose as SQL, losing baseline-correct tasks.

The three cases separate implementation, branch activation, and output
validity. Four hand-written controls with distinct paths yield
$6.67$\,pp headroom on $60$ tasks and $7.28$\,pp on $151$ items.
These post-inspection controls lack matched repeats and clones;
their $24/151$ development overlap is documented in
Appendix~\ref{app:phase1-provenance}. Worked traces follow in
Appendix~\ref{app:examples}.

\FloatBarrier
\subsection{Worked examples from the trace audit}\label{app:examples}

The six-of-twelve T1 count covers the full discovery population.
The separate trace audit follows five harnesses---the \bare{} baseline,
\react{}, a generated single-call candidate, a timeout-retry candidate,
and a multi-turn repair candidate---on ten tasks
(5 $\times$ 10). Below, one task illustrates these paths. Logged SQL
excerpts are capped at 120 characters; truncation is marked
\emph{[trunc.]}.

\paragraph{The task.} BIRD dev-set question 352 (\texttt{card\_games}
database; question text verbatim, the typo is the dataset's):
\begin{quote}\small\ttfamily
Calculate the percentage of the cards availabe in Chinese Simplified.
\end{quote}
Its BIRD hint:
\begin{quote}\small\ttfamily
'Chinese Simplified' is the language; percentage = Divide(Sum(id where
language = 'Chinese Simplified'), Count(id)) *100
\end{quote}
Gold SQL:
\begin{quote}\small\ttfamily
SELECT CAST(SUM(CASE WHEN T2.language = 'Chinese Simplified' THEN 1 ELSE 0
END) AS REAL) * 100 / COUNT(T1.id) FROM cards AS T1 INNER JOIN foreign\_data
AS T2 ON T1.uuid = T2.uuid
\end{quote}
The \bare{} harness answers:
\begin{quote}\small\ttfamily
SELECT SUM(CASE WHEN language = 'Chinese Simplified' THEN 1 ELSE 0 END) *
100.0 / COUNT(id) AS percentage FROM foreign\_data;
\end{quote}
On the released database, the gold query yields 8.7734 and the baseline
query yields 8.7728. The official scorer marks the baseline query wrong:
it counts directly over
\texttt{foreign\_data}, whereas the gold query counts after joining to
\texttt{cards}; the hint does not spell out that join.

\paragraph{\react{} (T2: untriggered repair).}
The first prompt is byte-identical to \bare{}'s. SQL executes successfully,
so the database-error retry does not fire. All 10 audited tasks follow
this path, with SQL execution but 10/10 final queries byte-identical to
\bare{}'s.

\paragraph{Single-call candidate (T1: absent repair).}
The docstring's \emph{conservative repair} specifies an output-token cap
and answer-only block. Execution consists of one model call followed
by SQL extraction. The cap is an environment default (2{,}048 versus
32{,}000 at runtime), has no harness control-flow implementation, and
never binds on these short answers. The worked task returns the same
wrong \texttt{foreign\_data} query, modulo whitespace, with one call
and zero SQL executions. Across all 10 tasks, SQL executions remain zero
and 6 final queries are byte-identical to \bare{}'s. The remaining
differences reflect a reworded prompt, without retrieval, self-check,
or repair.

\paragraph{Timeout-retry candidate (untriggered watchdog).}
A daemon-thread watchdog wraps the coder call and issues one hedged
duplicate after a 40\,s stall. No stall occurs in the audit:
SQL executions are zero and final SQL equals \bare{}'s on all 10 tasks.

\paragraph{Multi-turn repair candidate (T3: a broken multi-turn
path).} The one audited candidate implementing genuine execute-inspect-repair
returns the model's \emph{prose} as final SQL. On a task \bare{} answered
correctly (``State the alternative languages available for card named Annul
numbered 29.''), its logged final SQL begins, verbatim:
\begin{quote}\small\ttfamily
with "only a single read-only SELECT is allowed". The previous "query" was
actually just prose text (the model wrote rea\emph{[trunc.]})
\end{quote}
Under this repair re-prompt, the output contract fails again: asked to fix a query that ``was
actually just prose'', the frozen target responds with more prose reasoning,
no SQL block is found, and the reasoning text is submitted as the query. The
item flips from baseline-correct to harness-wrong.

\FloatBarrier

\FloatBarrier
\section{BIRD Admission-Policy Study}\label{app:phase2-protocol}

The $2{\times}2$ study uses $18$ builder--seed cells per arm
(six builders and three seeds); II-E is development-only.
Each cell has eight generation slots. Each slot allows three generation attempts and stops at first
admission without top-up. Two development databases are excluded.
Evaluation uses $1{,}169$ questions from nine databases and a pre-drawn
$400$-item core, with one disclosed $18$-item smoke-test exposure.
The official BIRD judge uses a $30$\,s timeout. Model-specific response
caching uses cross-process locks after correction of a concurrency defect.

The primary D$-$A contrast (forced/gated minus free/ungated headroom) pairs builder--seed cells with hierarchical
bootstrap and permutation inference; core analyses use Holm-corrected
factorial contrasts and K-matched sensitivity. Instrument fixes preceded
confirmatory outcomes. Subsequent analysis corrections share baseline
records and database/task resampling. Canonical records use
first-write-wins deduplication; conflicts and operational deviations
remain in the source audit. Generation costs vary across arms.

\begin{figure}[H]
\centering
\includegraphics[width=0.95\linewidth]{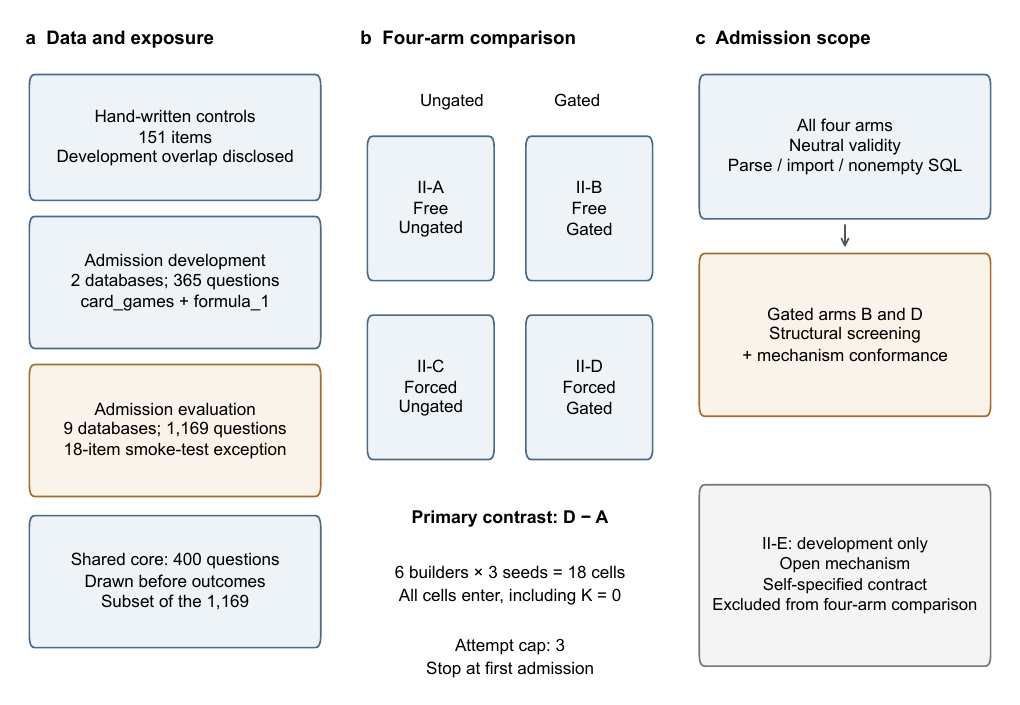}
\caption{\textbf{BIRD admission design.} Left: data splits and disclosed
development exposure. Middle: the $2{\times}2$ four-arm comparison with
a common attempt cap. Right: neutral validity and the implemented
structural-screening/conformance bundle in gated arms. The additional
open-mechanism arm II-E is development-only.}
\label{fig:design}
\end{figure}

\subsection{Design and primary contrasts}
\label{sec:f2-admission}
The four arms cross strategy forcing and gating:
II-A (free, ungated), II-B (free, gated), II-C (forced, ungated), and
II-D (forced, gated). Forced strategies come from a fixed set;
free builders select a declaration within the vocabulary in
Appendix~\ref{app:free-vocabulary}. Results below use the
corrected reanalysis of the pre-outcome specification.

\begin{table}[H]
\centering\small
\begin{tabular}{@{}lrr@{}}
\toprule
Metric & II-A & II-D \\
\midrule
Candidate count & 5.78 & 3.83 \\
Oracle accuracy (\%) & 76.14 & 75.70 \\
Best-fixed accuracy (\%) & 67.58 & 67.30 \\
Headroom (pp) & 8.56 & 8.40 \\
\midrule
D$-$A headroom (pp) & \multicolumn{2}{c}{$\RevisionPrimaryDelta$} \\
Unadjusted bootstrap 95\% CI & \multicolumn{2}{c}{$\RevisionPrimaryCI$} \\
One-sided sign-flip $p$ & \multicolumn{2}{c}{$\RevisionPrimaryP$} \\
\bottomrule
\end{tabular}

\caption{\textbf{BIRD admission: full-set headroom.} Means over
builder--seed cells on $1{,}169$ questions; \bare{} is included in
both the oracle and best-fixed comparators.}
\label{tab:primary}
\end{table}

The forced/gated minus free/ungated contrast is
$\RevisionPrimaryDelta$\,pp in baseline-inclusive oracle headroom,
with cell contrasts $\RevisionPrimaryRange$\,pp.
The implemented bundle does not improve observed mean headroom;
Table~\ref{tab:primary} reports uncertainty.

\begin{figure}[H]
\centering
\includegraphics[width=0.95\linewidth]{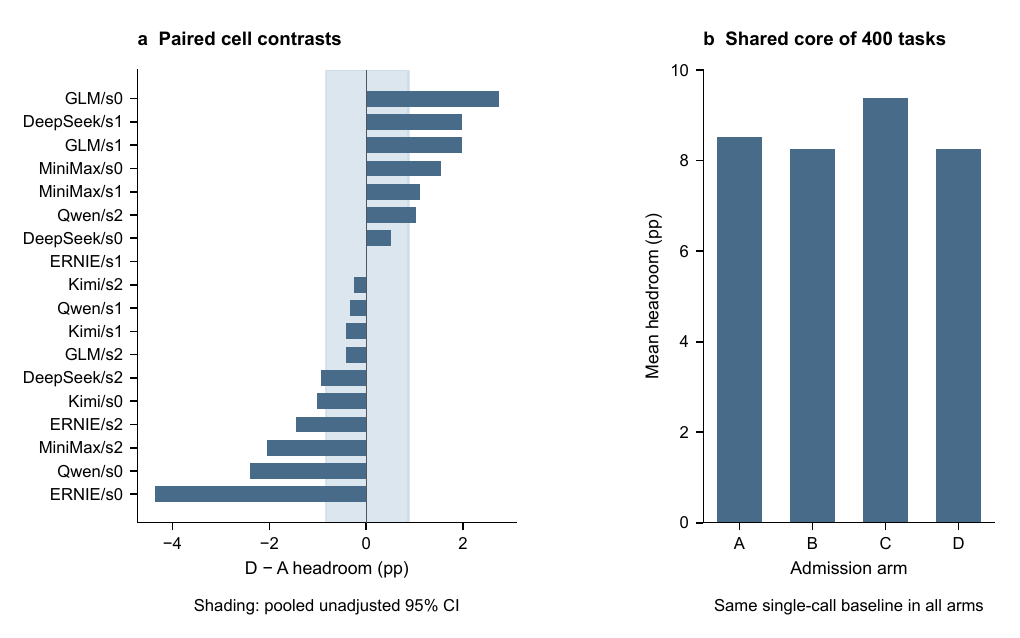}
\caption{\textbf{BIRD admission contrasts.} Left: full-set
D$-$A cell contrasts with the pooled 95\% interval. Right: four-arm
headroom on the shared core-$400$ tasks with a common \bare{} baseline.
Headroom and contrast axes show percentage points.}
\label{fig:results}
\end{figure}

On core-$400$, the gate contrast is $\RevisionGateDelta$\,pp,
with inconclusive strategy and interaction effects
(Table~\ref{tab:factorial}). The candidate-count-matched contrast is
$\RevisionKmatchedDelta$\,pp. II-E remains development-only.
The R2/R3 audits (Appendix~\ref{app:r2}) place these single-run
coverage measurements alongside $8.54\%$ R2 flips,
$14.5\%$ of baseline tasks changing across three runs, and a
$6.50$\,pp same-code oracle gap.

\begin{table}[H]
\centering\small
\begin{tabular}{@{}lrrrr@{}}
\toprule
Contrast & Effect (pp) & 95\% CI (pp) & Raw $p$ & Holm $p$ \\
\midrule
Gate & $-0.69$ & $[-1.56, +0.20]$ & 0.026 & 0.079 \\
Strategy & $+0.42$ & $[-0.32, +1.51]$ & 0.263 & 0.428 \\
Interaction & $-0.87$ & $[-2.83, +0.65]$ & 0.214 & 0.428 \\
\bottomrule
\end{tabular}

\caption{\textbf{Corrected core-$400$ factorial reanalysis.} Here A--D denote arm mean headrooms. Gate is
$[(B{-}A)+(D{-}C)]/2$, strategy $[(C{-}A)+(D{-}B)]/2$, interaction
$(D{-}C)-(B{-}A)$; raw and Holm-adjusted $p$-values.}
\label{tab:factorial}
\end{table}

\FloatBarrier
\Needspace{0.25\textheight}
\subsection{Same-core headroom decomposition}\label{app:decomp}
Table~\ref{tab:decomp} separates coverage, best-fixed accuracy, and
headroom using the same core tasks and baseline records in every arm.
\begin{table}[H]
\centering\footnotesize
\begin{tabular}{@{}lrrrrrr@{}}
\toprule
Arm & $K_{\rm cand}$ & $K_{\rm total}$ & Single-call (\%) & Best (\%) & Oracle (\%) & $H$ (pp) \\
\midrule
A & 5.78 & 6.78 & 63.25 & 66.00 & 74.51 & 8.51 \\
B & 6.11 & 7.11 & 63.25 & 66.12 & 74.39 & 8.26 \\
C & 5.61 & 6.61 & 63.25 & 65.64 & 75.01 & 9.38 \\
D & 3.83 & 4.83 & 63.25 & 65.65 & 73.90 & 8.25 \\
\bottomrule
\end{tabular}

\caption{\textbf{Same-core headroom decomposition.} All arms use identical
400 tasks and identical single-call outcomes. $K_{\rm cand}$ counts admitted candidates, excluding single-call, and
$K_{\rm total}=K_{\rm cand}+1$; counts and accuracies are cell means.
Headroom is oracle minus best-fixed accuracy, not realized routing gain.}
\label{tab:decomp}
\end{table}

\FloatBarrier
\subsection{Sensitivity, repeat audits, and recorded cost}
\paragraph{K-matched and R2 replays.}\label{app:r2}
Replays use deterministic record ordering, shared database/within-database
task draws, and seed resampling within fixed builders
(10{,}000 replicates). K-matching uses the smaller candidate count in
each A/B and C/D pair, exactly enumerates uniform subsets, retains
single-call, and reselects best-fixed within each subset and task draw.
It measures population-size sensitivity. R2, described below, retains
all 18 cells, including two singleton A subsets; its D$-$A changes from
\RevisionRtwoCached\ to \RevisionRtwoOff\,pp
(change \RevisionRtwoChange\,pp, 95\% CI \RevisionRtwoChangeCI).

R2 covers 70 harnesses on 400 tasks, taking the first two callable
admitted A/D candidates where available. Two cache-off acquisition groups
remain separate despite their duplicated repeat label, with program
identity checked against cached runs. Without matched single-call
acquisitions, both R2 conditions exclude that baseline; singleton
headroom is zero across the 18 retained pairs. Recorded flips are
\RevisionRtwoFlips/28{,}000 against cached outcomes
(\RevisionRtwoFlipPercent\%) and \RevisionRtwoRepeatFlips/14{,}400
between D cache-off passes (\RevisionRtwoRepeatPercent\%).
These describe observed instability: acquisition time and caching are
confounded, and the rates are not standard errors or effect-size floors.

\paragraph{R3 single-call-only repeat audit.}\label{app:r3}
Three independent cache-off runs evaluate only the single-call baseline
on core-400 with GLM-5.3-Flash at temperature 0. Correctness changes on
58/400 tasks (14.5\%); pairwise flips are 39/400, 40/400, and 37/400
(9.75\%, 10.0\%, 9.25\%). Treating runs as three pseudo-members,
one execution each, gives official-scorer oracle coverage $72.25\%$
and best-repeat accuracy $65.75\%$: a $6.50$\,pp same-code gap.
This records baseline execution variability at temperature 0
and motivates the repeat controls used in the main study.

\begin{table}[H]
\centering\small
\begin{tabular}{@{}lrr@{}}
\toprule
Sensitivity estimand & Estimate (pp) & 95\% CI (pp) \\
\midrule
K-matched B$-$A headroom & $-0.55$ & $[-1.53, +0.63]$ \\
K-matched D$-$C headroom & $-0.08$ & $[-1.26, +0.80]$ \\
K-matched gate mean & $-0.31$ & $[-1.07, +0.40]$ \\
\midrule
R2 subset D$-$A (cached) & $+0.42$ & $[-0.61, +1.48]$ \\
R2 subset D$-$A (cache-off) & $+0.81$ & $[-0.35, +1.84]$ \\
R2 change in subset D$-$A & $+0.39$ & $[-0.95, +1.63]$ \\
\bottomrule
\end{tabular}

\caption{\textbf{BIRD sensitivity on core-400.} CIs are unadjusted.
K-matched comparisons always retain single-call; R2 comparisons exclude single-call and
are conditional on the observed acquisition groups.}
\label{tab:sensitivity}
\end{table}

\paragraph{Recorded cost.}
The core records contain \RevisionLogicalCalls{} logical solver calls
across \RevisionCostRows{} candidate--task records
(Table~\ref{tab:cost}). These include cache-served requests and count
logical rather than billed requests; token usage is unavailable.
Generation attempts, admissions,
and execution calls are reported separately. Population cost averages
weight builder--seed cells equally, exclude single-call, and assign
zero calls to zero-candidate populations.

\begin{table}[H]
\centering\small
\begin{tabular}{@{}lrrr@{}}
\toprule
Arm & Candidate-task records & Logged solver calls & Calls/task/population \\
\midrule
A & 41,600 & 67,258 & 9.34 \\
B & 44,000 & 64,399 & 8.94 \\
C & 40,400 & 75,604 & 10.50 \\
D & 27,600 & 52,043 & 7.23 \\
\bottomrule
\end{tabular}

\caption{\textbf{BIRD generation and execution cost on core-400.}
Logged logical calls are not billable requests or token counts. The last
column averages candidate-population calls per task over cells.}
\label{tab:cost}
\end{table}

\FloatBarrier
\subsection{Admission scope and implementation}
\label{app:free-vocabulary}
Free-arm builders choose among repair, vote, twostage, and plain.
Vote requires multiple samples and selection, including but not limited
to three-way majority voting. Twostage consumes an artifact without a
prescribed response interface. Plain declares no control-flow mechanism
despite the prompt requesting a change. Thus strategy choice operates
within this four-declaration vocabulary.

The evaluated gate combines structural screening with conformance
verification; valid prompt-only transformations can fail the structural
screen. Development checks found false rejections, request-attribution
and response-interface defects, and unresolved judgments. Later
revisions produced no replacement benchmark outcomes, so results
describe the implemented bundle.

\paragraph{Generation interface.}\label{app:extraction}
Format-guard is one of eight fixed strategies, each assigned one slot
per builder--seed cell in both forced arms (C and D). It requests schema
fidelity and a final SQL code block with no extra text.
All $108$ format-guard attempts across its $36$ slots fail
neutral validity: $71$ logs report unterminated strings, $36$ no Python
fence, and one a provider timeout. Synthetic examples reproduce
truncation in the archived non-greedy fence extractor. Full builder
messages and exact historical extractor versions are unavailable,
limiting attribution of these upstream validity failures.

\FloatBarrier

\end{document}